\RequirePackage{fix-cm}
\documentclass[smallextended]{svjour3}       
\smartqed  
\usepackage{graphicx}
\usepackage{longtable} 
\usepackage{pdflscape} 

\begin{document}

\title{A Concept Specification and Abstraction-based \\ Semantic Representation}
\subtitle{Addressing the Barriers to Rule-based Machine Translation}


\author{Patrick C. Connor}


\institute{P. Connor \at
              73 Joeys Road, Mount Stewart, PE \\
              Tel.: +190-26-762524\\
              \email{patrick.c.connor@gmail.com}           
}

\date{Received: date / Accepted: date}

\maketitle

\begin{abstract}
Rule-based machine translation is more data efficient than the big data-based machine translation approaches, making it appropriate for languages with low bilingual corpus resources -- i.e., minority languages.  However, the rule-based approach has declined in popularity relative to its big data cousins primarily because of the extensive training and labour required to define the language rules.  To address this, we present a semantic representation that 1) treats all bits of meaning as individual concepts that 2) modify or further specify one another to build a network that relates entities in space and time.  Also, the representation can 3) encapsulate propositions and thereby define concepts in terms of other concepts, supporting the abstraction of underlying linguistic and ontological details.  These features afford an exact, yet intuitive semantic representation aimed at handling the great variety in language and reducing labour and training time.  The proposed natural language generation, parsing, and translation strategies are also amenable to probabilistic modeling and thus to learning the necessary rules from example data.
\end{abstract}

\section{Introduction}
There are a large number of minority languages in the world.  Although many of these are in danger of extinction \cite{Lewis_2009}, there are thousands of viable non-mainstream languages representing many tens of millions of people's native tongue.  While statistical machine learning and deep learning have been a boon to improving the quality of machine translation between the most common languages on the Internet, these approaches do not work well without the necessary substantial bilingual corpus, something often lacking for minority languages.  



Big data machine translation approaches are not going to be able to make implicit ideas/participants/features explicit in languages that require this to result in an accurate, clear, and natural-sounding translation. In rule-based machine translation (RBMT), there is room to include some contextualization to re-express gentive constructions, fill-out ellipsis, etc., where big data machine translation tends to perform a straight mapping between languages and is less likely to have mappings for all such situations.   Another advantage of RBMT over big data approaches is that there is no upper limit to the translation quality, since appropriate rules can model even anomalous or rarely used language that big data approaches would gloss over. 

 
RBMT can also take advantage of an interlingua, a carefully designed intermediate or ``hub'' language.  In such a situation, translation between \emph{any} two languages (including minority languages) requires only that they be defined in terms of the interlingua.  Translation proceeds by first translating from the source language into the interlingua and then from the interlingua into the receptor language. The key value of using an interlingua is that it only requires one translation model per supported language.  In contrast, modern statistical machine learning approaches technically need a bilingual corpus for each translation pair, although they sometimes work through one or more intermediate languages such as English that are not designed as an interlingua to find a path with enough bilingual corpus to generate reasonable results. 

In spite of its potential benefits, RBMT is not the engine behind commonly used translation products. RBMT approaches are thought to be less fluent~(e.g., \cite{Tyers_2013}, \cite{SreelekhaEtAl_2018}) than statistical approaches, perhaps because achieving fluency would involve a great deal more rules and complexity than translation \emph{adequacy} requires.  RBMT can be practical for adjusting a text for a similar dialect~\cite{ForcadaEtAl_2011}, but has been very challenging otherwise.  The trouble is that crossing the morphosyntactic and semantic domains of two disparate languages is very complicated and rule interactions can become unwieldy. To further complicate matters, ``Whatever the power of representation of a (meta) language is, there exist some linguistic phenomena requiring exceptional processing..."~\cite{SenellartEtAl_2001} (i.e., special cases).  However, \emph{the greatest barrier to RBMT appears to be the large amount of training and human effort required to build the language rules.}  


At the heart of solving these RBMT challenges, we believe, is a \emph{user-friendly and statistically tractable} semantic representation.  As will be discussed in Section~\ref{sec-relatedwork}, existing related representations are capable of translation and/or NLG, but they require the manual building of new language models and the encoding of content in terms of their interlingua, which requires substantial human training, time, and effort.  Also, because of their complexity, it is doubtful that serious data-driven statistical models could be effectively employed to help develop new language models.  Thus we propose a new semantic representation for RBMT with the following goals:
\begin{itemize}
\item provide the simplest, most accessible way to represent and model meaning (for humans and computers) and leave room for users to easily tailor the representation as required
\item have a natural affinity for natural language generation (realisation), parsing, and statistical tractability
\item span the representation of meaning from the individual concept to mid- or high-level discourse relations
\item be expressive enough to encode all aspects of meaning expressed in the surface text of any language
\item prove consistent, unambiguous, and canonical:
\begin{itemize}
   \item given a sentence with no inherent ambiguity, there should be exactly one semantic representation that is also unambiguous
   \item different sentences should share the same semantic representation when they share exactly the same underlying concepts, which are related to one another in the same way. 
\end{itemize}
\end{itemize}
   
The rest of this paper is organized as follows.  Section~\ref{sec_exintro} introduces the semantic representation with rationale for the principle features of the approach. Using numerous examples, Section~\ref{sec_guidlines} provides guidelines for how to represent a variety of linguistic phenomena in terms of the approach.  Section~\ref{sec-genparse} describes a general strategy for transduction and translation and section~\ref{sec-relatedwork} relates three thematic role-based representations, which are among the most similar semantic representations to the present approach, each employing their own interlingua.

\section{An Example-based Introduction}
\label{sec_exintro}
To achieve the above-named goals, our semantic representation is defined by several key principles:
\begin{itemize}
\item separation:  all divisible elements of meaning (morphemes+) in a sentence are expressed as individual concepts
\item specification:  concepts further specify or narrow the meaing of other concepts to form networks of meaning that relate entities to one another.
\item encapsulation:  individual conceptual networks can be encapsulated and treated as a single concept.
\end{itemize}

\begin{figure}[h]
\centering
\includegraphics[width=\textwidth]{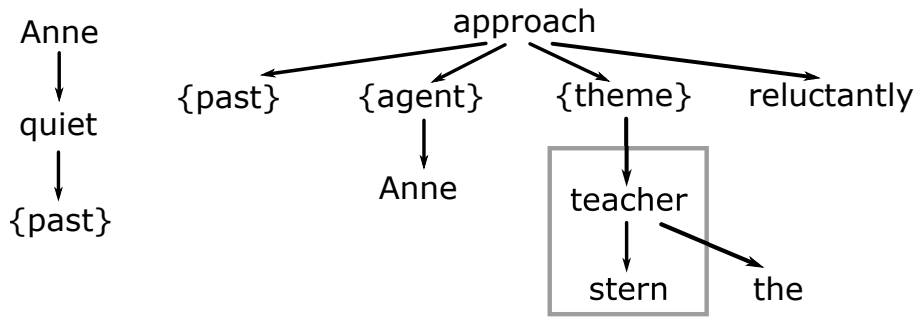}
\caption{Tree diagrams for, ``Anne was quiet," and, ``Reluctantly, Anne approached the stern teacher."  Arrows point to concepts that further specify or narrow their parent's meaning.}
\label{fig:treediags}
\end{figure}

Our representation takes the form of a tree. Figure~\ref{fig:treediags} provides two examples.  In them, each node is a concept and each arrow points from a concept being further specified to the concept providing the additional detail. Although rare, the tree can have more than one root, and will often have many leaf nodes, which usually specify/narrow only one concept, but sometimes specify more than one. The trees can generally be written into a single line of text that we refer to here as tree-line notation, which provides an easy mode of data entry.  Let's illustrate the three key principles of our approach with these two network formats using a running narrative example.

Across a wide variety of linguistic phenomena we find one concept ``modifying'' another (e.g., ``linguistic'' modifies ``phenomena'' in this sentence).  More broadly, concepts tend to narrow or further \emph{specify} the details of other concepts.  In a semantic representation, it is important to have a consistent way of relating concepts for network building and usage.   Specification provides a natural, intuitive rule by which to relate concepts.  An important feature of Universal Conceptual Cognitive Annotation~\cite{AbendRappoport_2013} is that it requires less training of annotators than do other approaches, involving only a dozen semantic relations to learn for constructing graphs.  In the present approach, we pursue this goal by embedding the relationships entirely in the graph structure using specification.  When uncertain about the arrangement between two concepts, the question, ``Which concept specifies or narrows the meaning of the other," can help clarify.  

\begin{equation}
``Anne\: was\: quiet."
\end{equation}
\begin{equation}
Anne > quiet > \{past\}
\end{equation}

In this stative sentence, the attribute concept ``quiet" further describes or specifies the entity (noun) concept ``Anne", and ``quiet" is further specified as a ``\{past\}" tense attribute.  Note that a convention we use is to surround a concept in ``\{\}" when it has no natural stem of its own in the language being modeled (e.g., tense, plurality, tone, etc.) but rather affects the expression of other tems.  For example, we view tense as an attribute specifier or a verb specifier rather than as a distinct verb (see Section~\ref{sec-copula} on tense and copula for more details).  



\begin{eqnarray}
``Reluctantly,\: Anne\: approached\: the\: stern\: teacher."
\end{eqnarray}
\begin{eqnarray}
\nonumber approach > [\{past\}, \{agent\} > Anne, \{theme\} > \\
(teacher > stern) > the, reluctantly]
\end{eqnarray}

The entity ``teacher" is described as being ``stern", the determiner ``the" specifies a certain stern teacher, and the adverb ``reluctantly" describes how the ``approach" was made.  Event concepts (verbs) in propositions will tend to form the roots of trees and entities will form the branches.  Adjectives, adverbs, and determiners often make up the leaves. 

In this event proposition, the \{agent\} and \{theme\} role concepts are introduced.  These concepts are needed to specify which person takes which role.  They are used instead of the often matching terms ``subject'' and ``object'' because these specify a syntactic role rather than a semantic role (sometimes the agent is the object in a sentence).  To keep the representation simple,\{agent\} and \{theme\} are represented as concepts rather than as a separate class of ``relations'' \cite{BealeEtAl_2005}, \cite{Allman_2010}, \cite{UchidaEtAl_2006} or function ``slots" \cite{MitamuraEtAl_1991}.  If the roles were not specified, they would face the same risk as approaches that represent a proposition in a vectorized format, also known as sentence embeddings.  To maintain a constant length vector in such an approach, either there must be regions of the vector devoted to each possible role or else strongly similar concepts with different roles (e.g., ``Anne'' and the ``teacher'') must mingle (e.g., \cite{AroraEtAl_2016}) and risk confusion.  

Implicitly, a semantic representation should strip away syntactic or morphological features, leaving the meaning behind. To achieve this, every bit of meaning encoded by syntax or inflectional morphology is converted into a separate concept.  In our example, the verb ``approached" is decomposed into its constituent stem ``approach" and tense ``\{past\}" elements.  The benefit is a common basis for simpler transfer to other languages, which, for example, may not express tense by inflecting the verb.  Indeed, handling details of morphology is often left out of a semantic representation.  This has a relatively minor impact in English, but in languages where subject and/or object can be expressed via bounded morphemes (e.g., Orizaba Aztec~\cite{ElsonPickett_1983}) it will be more important. Implied concepts are also included in our approach by appropriately inserting in the network the implied concept and specifying it with the \{implied\} concept (see Table~\ref{tab_examples} for examples).  Also in the present approach, all elements of meaning are expressed as \emph{concepts only}.  In contrast, first-order logic, for example, employs various classes to embody elements of meaning including terms, predicates, functions, and quantifiers. This can lead to awkwardness and language expressibility difficulties (e.g., quantifiers cannot be functionalized \cite{Schubert_2015}). Using a single meaning element class avoids the complication of crossing the boundaries between classes of meaning. Symbols of logic rendered as concepts along with their conceptual arguments can still be identified and used for conducting inferences in this approach as needed.

Using square brackets in the tree-line notation enables the example concept ``approach" to be specified by a list of concepts.  Specification by multiple concepts has no formal ordering, no required set of specifying concept arguments, and technically no limit to the number or variety of concepts that may be included.  In contrast to some approaches (e.g., FrameNet~\cite{FillmoreEtAl_2003}), this means that users of a language model need not know or look up any particulars (e.g., functional arguments, class, etc.) of any given concept before invoking it, except perhaps for differentiating between homographs (two or more concepts with the same spelling).  

The ability to encapsulate and treat a proposition as a new \emph{concept} is powerful, achieving several important functions.   In the previous example, it allowed us to build up a more descriptive entity before refining it further. It would allow distinguishing between the ``stern teacher" in the narrative and possibly a different teacher in the vicinity. It provides a way of ordering of operations, in support of unambiguity.  In the noun phrase, ``all silk clothing'', it allows us to encode ``(all silk) clothing'' and ``all (silk clothing)'' differently. This is also a feature of Combinatory Categorial Grammar~\cite{Steedman_1987} where the semantic relationships have an explicit order of operations or nested groupings based on the order in which combinatory rules are applied.

\begin{eqnarray}
\nonumber ``The\: teacher\: softened,\: however,\: because\: \\ 
Anne\: was\: crying." 
\end{eqnarray}
\begin{eqnarray}
\nonumber (soften > [\{past\}, \{agent\} > teacher > the]) > \\
because > (cry > [\{past cont.\}, \{agent\} > Anne])
\end{eqnarray}

Here, encapsulation has a different function, where two propositions enclosed by parentheses are related via the ``because" concept, capturing the cause and effect nature between two separate propositions.  Encapsulation can lead to even higher, discourse-level organization of meaning, which might prove helpful in applications like text summarization.  The word ``however" in the current example could be related back to the teacher's sternness in a similar fashion. 

All encapsulations discussed so far only temporarily house a small network of concepts.  But if a particular network of concepts were used frequently enough, it would seem worthwhile to label it and refer to it thereafter by name.  Defining concepts in terms of other concepts is an important feature of our approach.  For example, if we wanted to define an ``Anne'' as an imaginative girl, we could represent this as,

\begin{eqnarray}
Anne = girl > imaginative,   
\end{eqnarray}

where $girl = human > [young, female]$, which tells us that ultimately an ``Anne'' is human.  This capability provides a convenient basis for abstraction of underlying linguistic and ontological details. Suppose we were to define forks, fly swatters, and staplers as ``tools'' with associated specifications.  Then, when grammar treats tools in a certain way, these objects would be treated similarly without having to tag them with each use, as is done in other representations.  Later, suppose you chose to define a pitchfork in terms of a fork.  It would then automatically inherit its status as a tool.  Another benefit of this abstraction is that after all the definitions are in place, users and algorithms that know nothing about the language definitions/rules will have the same benefits and will not have had to learn to label tools.  

In relation to the gamut of semantic representations \cite{Schubert_2015}, \cite{AbendRappoport_2017}, the present representation falls somewhere between sentence diagramming and a semantic network having logical form. In sentence diagramming (e.g., Reed-Kellogg \cite{ReedKellogg_1896}), all of the words of the sentence must be used and be fit into the framework.  Our approach naturally leans toward the use of all underlying concepts in the sentence, even the most minute, but does not force all words to be included, since some words serve only syntactic purposes and do not add to the meaning (e.g., Boys are in the yard vs. \emph{There} are boys in the yard).  Logical form semantic networks like Abstract Meaning Representation \cite{BanarescuEtAl_2013} drop pronouns and instead link their actions to the entities they reference, building a graph of logical interactions between entities.  While our approach is not averse to the use of additional processing to produce similarly abstract networks, it natively maintains all of the underlying concepts in the surface structure, including pronouns, articles, tense etc. to provide the most direct and complete interpretation of the surface structure's meaning.  This will help preserve the same meaning in translation to another language, with as little disturbance to the original content as possible.

This approach is very similar to dependency-based representations.  However, it differs in a few ways.  Dependency-based representations include all the words of a sentence in their graph structure, whereas the present approach does not but will instead include implied concepts.   Dependency-based approaches tend not to break a morphologically inflected rule into its constituent morphemes, whereas the present approach does and adds a few case roles into the graph as well.  There is hierarchy in a dependency-based grammar, but there is no way to indicate that one dependency should be modify its parent before another, whereas the present approach can do so.  Finally, the present approach aims for every child word to further specify its parent whereas dependency grammars may sometimes use this rule of thumb but do not strive to do so wherever possible.

A collection of novel properties of this approach, which will be elaborated upon in the sections that follow, are:
\begin{itemize}
\item all elements of meaning, including case roles, are captured in concepts rather than as ``features'', ``tags'', or ``relations'', to streamline and simplify the representation
\item encapsulation of subnetworks of concepts with specified a head word and ``foot'' word, to hierarchically organize and arrange the precedence of specifying concepts 
\item a highly user-friendly lexical and rule definition approach, to make usage very intuitive
\item the unification of syntactic and morphological rules, to streamline rule creation
\item inclusion of ontological information within the semantic network, to serve in rule selection
\item rule selection for parsing and realisation based on ontological similarity between network concepts and rule concepts, to allow users to make use of rules without learning lexical or semantic categories.
\end{itemize}

\section{Guidelines for use}
\label{sec_guidlines}
In the previous section, examples were provided to explain the primary mechanisms of the representation.  Here, we propose ways of using our approach to interpret the meaning of certain types of concepts and surface structures, with examples provided in Table~\ref{tab_examples} that match the discussion in the subsections that follow.  Although our examples\footnote{Some of our examples come from Wikipedia.} are kept short for clarity, sentences of any length can be represented.  Our goal is to adhere to natural intuitions about how concepts are related, even when this not straightforward, and to generate similar representations for sentences with similar meaning even when the syntax is different.  

\small
\begin{landscape}
\begin{longtable}{lll}
\caption{Various aspects of language expressed in this semantic representation.} \\
\hline\noalign{\smallskip}
\label{tab_examples} 
Type & Example Sentence & Tree-line Notation \\ \hline
\noalign{\smallskip}\hline\noalign{\smallskip}
Adjectives     & The dress is green.  & (dress $>$ the) $>$ green $>$\{present\}  \\
Adverbs       & Quickly, he ran home.   &  run $>$ [\{past\}, \{agent\} $>$ he, home, quickly]  \\
Determiners     & John lifted a rock.   & lift $>$ [\{past\}, \{agent\} $>$ John, \{theme\} $>$ rock $>$ a]  \\
                       &  He found this worm.         &  find $>$ [\{past\}, \{agent\} $>$ he, \{theme\} $>$ worm $>$ this]  \\
Prepositions         & The ball is under the table.  &  (ball $>$ the) $>$ (under $>$ table $>$ the) $>$ \{present\}         \\
  				    &  It flew through the window.   & fly $>$ [\{past\}, \{agent\} $>$ it, through $>$ window $>$ the]       \\
Emphasis     &  Holy cow!    &  holy cow $>$ \{!\}         \\
                  &  *She* did not eat it.    & eat $>$ [\{past\}, \{agent\} $>$ she $>$ not, \{theme\} $>$ it]         \\
Intensive pronouns     & He himself bought the car.  & buy $>$ [\{past\}, \{agent\} $>$ he $>$ \{emphasis\}, \{theme\} $>$ car $>$ the]  \\
Numbers          & John will order two hamburgers.     & order $>$ [\{future\}, \{agent\} $>$ John, \{theme\} $>$ hamburger $>$ two        \\
Plurality            & Mary will pick up the eggs.   & pick up $>$ [\{future\}, \{agent\} $>$ Mary, \{theme\} $>$ (egg $>$ \{plural\}) $>$ the]     \\
Logic concepts     & All silk clothing is comfortable.  & ((clothing $>$ silk) $>$ all) $>$ comfortable $>$ \{present\} \\
                         & Either he or Mary will pay.   & pay $>$ [\{future\}, \{agent\} $>$ either or $>$ [he, Mary]]   \\
		              & I did not go.  & (go $>$ [\{past\}, \{agent\} $>$ I]) $>$ not \\
Modals               & She can go.   &  (go $>$ [\{present\}, \{agent\} $>$ she]) $>$ can         \\
			         & He must not stay.  & (stay $>$ [\{present\}, \{agent\} $>$ he]) $>$ must $>$ not   \\
Questions     & Did it break the window?         & (break $>$ [\{past\}, \{agent\} $>$ it, \{theme\} $>$ window $>$ the]) $>$ \{?\}    \\
                  &  Who threw the ball?               &  throw $>$ [\{past\}, \{agent\} $>$ who $>$ \{?\}, \{theme\} $>$ ball $>$ the]          \\
Tense            &  The rain washed the truck.  &  wash $>$ [\{past\}, \{agent\} $>$ rain $>$ the, \{theme\} $>$ truck $>$ the]   \\
Copula                       &  Fred is happy.               &    Fred $>$ happy $>$ \{present\}         \\
                               &  Fred seems happy.        &     Fred $>$ happy $>$ seem $>$ \{present\}       \\
                               &  It seems that Fred is happy.        &     Fred $>$ happy $>$ seem $>$ \{present\}       \\
Existential Clause       &  There are boys in the yard.       &  ((boy $>$ \{plural\}) $>$ (in $>$ yard $>$ the)) $>$ \{present\}   \\
Reification                 &  Fred is the plumber.       &   \{re\} $>$ [Fred, (plumber $>$ the)] $>$ \{present\}       \\
                               &  The plumber is Fred.       &   \{re\} $>$ [Fred, (plumber $>$ the)] $>$ \{present\}       \\
Named Entity            &   Canada is not always cold.   &   Canada $>$ (cold $>$ always $>$ not) $>$ \{present\}      \\
Pronoun                   &  We polar dip in January.       &   polar dip $>$ [\{agent\} $>$ we, in $>$ January]       \\
Noun phrase             &  The dog that ate the  &   bark $>$ [\{past\}, \{agent\} $>$ dog $>$ (eat $>$  [\{past\},  \\
              	               &  \,\, peanut butter barked happily.  &   $>>$\{agent\}, \{theme\} $>$(butter $>$ peanut) $>$ the]), happily]      \\
                               &  Who does Carl believe that  &   believe $>$ [\{present\}, \{agent\} $>$ Carl, \{theme\} $>$ \\
              	               &  \,\,Bob knows that Mary likes? & (know $>$ [\{present\}, \{agent\} $>$ Bob, \{theme\} $>$  \\
              	               &              	                    &  $>>$(like $>$ [\{present\}, \{agent\} $>$ Mary, \{theme\} $>$ $>>$ who $>$ \{?\}])])]      \\
Participles       &   The trembling bird flew away.   &    fly $>$ [\{past\}, away, \{agent\} $>$ (bird $>$ (tremble $>$ $>>$\{agent\})) $>$ the]   \\
Gerunds                    & Swimming is fun.  &  (swim $>$ \{agent\} $>$ I $>$ \{implied\}) $>$ fun $>$ \{present\}       \\
Multi-Object Verbs       & John traded Mary his marble &    trade $>$ [\{past\}, \{agent\} $>$ John, \{recipient\} $>$ Mary, \\
                               &  \,\, for her ribbon.                           &   \{object 1\} $>$ marble $>$ his, \{object 2\} $>$ ribbon $>$ her]      \\
Linking Verbs             &  That chicken tastes good.  &  (chicken $>$ that) $>$ ((taste $>$ good) $>$ \{present\})       \\
                               &  It smells like fish.  &  it $>$ smell $>$ (like $>$ fish) $>$ \{present\}       \\
Verbs Followed           & I want to go home.        &   want $>$ [\{present\}, \{agent\} $>$ I, \{theme\} $>$ (go $>$ [\{agent\} $>$ I $>$ \{implied\}, home])]      \\
 by Infinitives             & She tried to tell you.  &     try $>$ [\{past\}, \{agent\} $>$ she, \{theme\} $>$     \\
                               &                        & (tell $>$ [\{agent\} $>$ she $>$ \{implied\}, \{theme\} $>$ you])]        \\
Ergative verbs           & The sun melted the ice.    & melt $>$ [\{past\}, \{agent\} $>$ sun $>$ the, \{theme\} $>$ ice $>$ the]        \\
                               & The ice was melted.   &   melt $>$ [\{past\}, \{theme\} $>$ ice $>$ the]      \\
                               & The ice melted.        &   melt $>$ [\{past\}, \{theme\} $>$ ice $>$ the]      \\
Possessives                & John's dog is hungry.  &  (dog $>$ (have $>$ [$<<$\{agent\}, $>>$\{theme\}]) $>$ John) $>$ hungry $>$ \{present\}      \\
Voice/Topicalization    & The truck was driven by John.   &  drive $>$ [\{past\}, \{agent\} $>$ John, \{theme\} $>$ [truck $>$ the, \{topic\}]]       \\
Prepositional phrase   &  John fell through the   &  fall $>$ [\{past\}, \{agent\} $>$ John, through $>$ \\
                               &  \,\,  rotting floor.                                                &    (floor $>$ (rot $>$ [\{present continuous\}, $>>$\{agent\}])) $>$ the]       \\
Similie and Degree    & She ran like a scared rabbit   &  run $>$ [\{past\}, \{agent\} $>$ she, like $>$        \\
                               &         &  (run $>$ [\{agent\} $>$ (rabbit $>$ (scare $>$ [\{past\}, $>>$\{theme\}])) $>$ a, $>>$\{how\}])]       \\
                               & The brown dog is bigger  &   ((dog $>$ brown) $>$ the) $>$ (\{more than\} $>$ big $>$ \{present\}) $>$ \\
                               &  \,\, than the black dog.       &  (dog $>$ black) $>$ the        \\
Cause and Effect       & If John confesses then Bill &   (Bill $>$ angry $>$ \{future\}) $>$ if $>$ (confess [\{future\}, \{agent\} $>$ John])      \\
                                  & \,\, will be angry.  & \\  
Contrast                   & She hid but I found her.         &   (hide $>$ [\{past\}, \{agent\} $>$ she]) $>$ but $>$  \\
                               &                                              &     (find $>$ [\{past\}, \{agent\} $>$ I, \{theme\} $>$ she])       \\
Groups                     & I like her but she   &   but $>$ [like $>$ [\{present\}, \{agent\} $>$ I, \{theme\} $>$ she],       \\
                               &   \,\, doesn't like me.  &  like $>$ [\{present\}, \{agent\} $>$ she, \{theme\} $>$ I, not]]        \\
                               & Either she stays or &  either...or $>$ [stay $>$ [\{present\}, \{agent\} $>$ she],     \\
                               &  \,\, she goes away.     &  go $>$ [\{present\}, \{agent\} $>$ she, away]] \\
Quotation		&  She replied, ``I will leave.''        &  reply $>$ [\{past\}, \{agent\} $>$ she, \{theme\} $>$ \{quote\} $>$ \\
					    &									 &	 (leave $>$ [\{future\}, \{agent\} $>$ I])] \\
Sequence                 &  She was frightened.  She ran.    &   (frighten $>$ [\{past\}, \{theme\} $>$ she]) $>$ \{seq\} $>$  (run $>$ [\{past\}, \{agent\} $>$ she])       \\
Nonsense   & Colourless green ideas &  sleep $>$ [\{agent\} $>$  ((idea $>$ \{plural\}) $>$ green) $>$ colourless, furiously]\\
                   &  \,\, sleep furiously. &   \\ \hline
\noalign{\smallskip}\hline
\end{longtable}
\end{landscape}
\normalsize

\subsection{Straightforward Concept Specification}
Many standard aspects of language can be represented simply as one concept specifying or narrowing the meaning of another.  Adjectives and adverbs, for instance, describe how an event was performed or specify an attribute of an entity.  Tense narrows the details of an event or an attribute concept by specifying timing.  Determiners further specify the instance of the concept  they refer to.  Prepositions and prepositional phrases further specify where in space or time an entity is or an event takes place.  Emphasis via bold font, italics, exclamation marks, etc. adds to meaning and can also be tracked as a specifying, stemless concept.  Such emphasis sometimes serves to identify \emph{which} concept(s) should be further specified by others.  Intensive pronouns add emphasis in a different way but are represented in the same way.  Numbers are simply a further specification of an entity, as is plurality.  Logic concepts and modals are also simple specifications on concepts.  Naturally, we consider questions to be substantially different from affirmative statements or commands.  This is certainly true for how people respond, but in terms of representing their meaning, our approach employs another simple specification by applying the \{?\} concept to the unknown concept.  In true-or-false questions it is usually applied to an entire proposition, but could also be applied to a particular concept within if an appropriate emphasis were applied.  Wh- questions involve specifying the appropriate wh- concept with the question concept.  The notion of concepts further specifying one another is a powerful one that unifies the representation of a large number of disparate linguistic phenomena.  

\subsection{Tense, Copula, and Reification}
\label{sec-copula}
So far, we have shown tense being used to further specify events and even the attributes of entities.  Events are naturally specified by tense, expressing roughly when the action occurs.  When applied to entities, tense specifies the timing of another specifying concept.  This steps into the realm of copula, where in English the verb ``to be" (often expressed by ``is" and ``was") is not treated as an event but rather as a means of expressing the tense or timing of an attribute specifying a concept.  By convention, tense concepts in our representation are always surrounded with ``\{ \}" and make up a large portion of the stemless concepts.  As an aside, the linguist has the freedom to adjust these stemless concepts to suit their purposes.  For example, it would make sense to abbreviate the most commonly used of these to make data entry more efficient.  Also some languages will need a different set of them (e.g., different tenses).  We will look at how to represent most semi-copulas or linking verbs in our later discussion of event concepts.  However the concepts ``seem" and ``appear" behave somewhat differently, like a modal.  Existential clauses sometimes do and sometimes do not behave as a copula.  In the sentence, ``There is a God," the standard interpretation is, ``A God exists."  However, in the sentence, ``There are boys in the yard," the interpretation is not that ``boys exist ... in the yard", but that an instance or reification of ``boys" has the prepositional attribute of being ``in the yard".  

Reification is the process of making real or creating an instance of a concept.  For example, when we talk of ``a dress" we are not talking of the general concept ``dress" but a particular dress.  Most of the time, entities referred to in text are reified.  When beneficial, one way to represent reified concepts is to have that concept specify a stemless reification concept (e.g., ``\{re\} $>$ dress $>$ a"). For clarity, and because in a computerized  implementation copies or instances of the concepts will be used in the semantic networks, we do not normally include this.  Technically it should not be contained in the underlying definition of concepts since those are not reifications.  It comes in handy, though, when two or more reified entities refer back to the same underlying entity.  Consider the example, ``Fred is the plumber."  Both ``Fred" and ``the plumber" are separately reified entities, but the copula used here tells us that they refer to the same entity and could be represented as ``\{re\} $>$ [Fred, plumber $>$ the] $>$ \{present\}".  

\subsection{Specialized Entities or Nouns}
Named entities, as well as pronouns can be specified in the same way as other entities.  Named entities such as ``Canada" or ``South America" could possibly be reified by definition since there is only one real-world instance of them.  

\begin{figure}[h]
\centering
\includegraphics[width=0.65\textwidth]{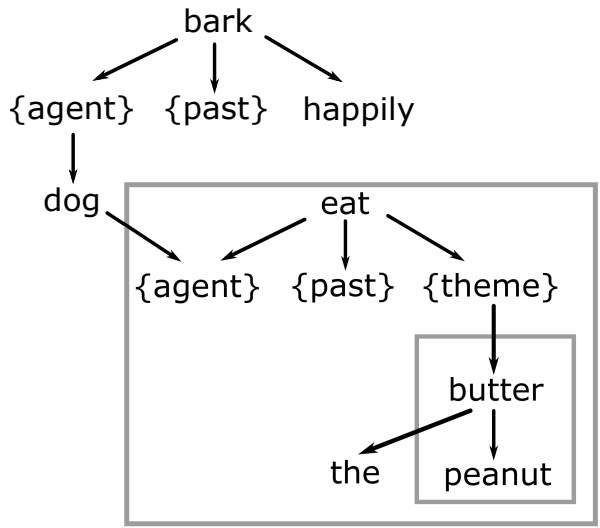}
\caption{Tree diagram for ``The dog that ate the peanut butter barked happily."  The entity ``dog" is specified by the agent of a relative clause.}
\label{fig:relclause}
\end{figure}

A different, but common, way of specifying a noun is with reference to a previous event.  For example, ``The dog that ate the peanut butter barked happily."  In this sentence, the dog appears as the agent of two events.  One of the propositions specifies which dog is being talked about in the other proposition. The semantic representation of this sentence, having a relative clause and diagrammed in Figure~\ref{fig:relclause}, has the tree-line notation of 

\begin{eqnarray}
\nonumber bark > [\{past\}, \{agent\} > dog > (eat > \\
\nonumber [\{past\}, >>\{agent\}, \{theme\} >\\
(butter > peanut) > the]), happily]
\end{eqnarray}

In our conception, the barking dog is further specified as the agent of another proposition. In tree-line notation, we make the connection through an encapsulation boundary ``('' with the prefix ``$>>$". Unless so indicated, the head concept within an encapsulated proposition defaults to the left-most concept in parentheses in the tree-line notation, which is usually the verb in an event proposition.  When we want to designate one of the encapsulated concepts to be further specified by a concept on the outside, we would add the prefix ``$<<$".  In some cases this is useful, as in, ``She loved the cat that caught the rat that found the cheese."  Some definitions can make use of this mechanism as well.  We can represent, ``Mary is a beautiful singer," as  Mary $>$ ($>>$(sing  $>$ [$>>$\{agent\}, beautiful])  $>$ a) $>$ \{present\}.  Here, we specify Mary as the agent of ``sings beautifully''.  Here, we used two ``$>>$" prefixes in the same tree-line structure because we have to drill down into two encapsulations to reach the specifying concept.  We can see several situations that make use of the $>>$ prefix in Table~\ref{sec_guidlines}.  For example, participles are verbs that behave like an adjective.  Instead they can be seen as suggesting that the entity they specify is rather specified as the agent of the associated verb.  For example, ``singing Mary'' would be represented as Mary $>$ (sing $>$ [\{present cont.\}, $>>$ \{agent\}]). In English, many words ending in -able can be seen as encapsulated propositions.  For example, ``bendable metal" can be represented as metal $>$ (bend $>$ [$>>$\{agent\}, can]), which is the same representation for, ``metal that can bend''.  An encapsulated proposition can also behave as an entity.  For example, a gerund treats a verb as a noun (e.g. ``Swimming is fun."). 

\subsection{Specialized Events or Verbs}
Most of the events we have dealt with so far have at most one object.  However, some events have more than one object.  For example, the verb ``to trade" has four roles that must be identified:  agent, recipient, object 1, object 2.  In such circumstances the linguist can create new stemless concepts like the \{agent\} and \{theme\} to denote these roles in the semantic representations.  Our approach only employs such thematic role concepts when there are two or more entities (e.g., John and Mary) that could sensibly fill any of two or more roles (e.g., the \{agent\} and \{theme\} roles).  Necessary use of explicitly labeled roles beyond agent, theme, recipient, and instrument should be rare, especially since agent and theme could each be used in place of a several other related roles. 

There are some categories of verbs that do not behave like standard events.  Linking verbs (e.g., tastes, feels, looks) appear in place of the copula ``to be" in English and reflect the state of the entity they specify rather than indicate an event between entities.  There is a group of verbs that are often followed by infinitives (e.g., ``I want to go home'', ``She tried to tell you'').  These can be seen as an event with a \{theme\} that is itself an encapsulated proposition.  Ergative verbs are such that when you drop the subject (making them intransitive), the object becomes a subject (e.g., The sun melted the ice vs. The ice melted (intransitive form)) yet remains the \{theme\}. Light verbs such as make, take, give, do, etc., are treated in the same way as normal verbs.  If it is the desire of the linguist to simplify the meaning structure by removing them, this can be done.  However, in principle, our approach allows users to stick as closely as possible to the surface text, only dropping words that do not add to meaning at all. To reiterate, our representation allows the linguist to use the provided conventions for how to handle the different types of syntactic phenomena or to adjust them as they wish.  Consistency within a language model, however, will be valuable for natural language generation and parsing.

Possessives can be represented in terms of the verb ``to have".  The possessor becomes the agent and the thing possessed is specified by the \{theme\} of ``to have".  In ``John's dog is hungry", we get the representation, ``(dog $>$ (have $>$ [$<<$\{agent\}, $>>$\{theme\}]) $>$ John) $>$ hungry $>$ \{present\}".  If beforehand we define ``\{have\} =  (have $>$ [$<<$\{agent\}, $>>$\{theme\}])", the representation is reduced to ``(dog $>$ \{have\} $>$ John) $>$ hungry $>$ \{present\}".  As an aside, we can use the same approach to define the composition of entities (e.g., bird $=$ organism $>$ \{have\} $>$[wing $>$ two, feather $>$ \{plural\}]).  It is tempting to just use the \{have\} concept without bothering to define it, but if this were our policy regarding definitions, we might miss out on the use of data examples that inform generation and parsing according to the underlying structure within the definition. 

Voice and topicalization make certain concepts in a sentence the focus.  This could be simply noted using a \{topic\} stemless concept to specify the focal entity.  In a multi-proposition sentence, an entire proposition as well as an entity within it may need such a marker. 


\subsection{Multiple Propositions and Beyond}
Many sentences contain two or more propositions, connecting to one another by different means.  Some prepositional phrases behave in this way and use a preposition or connecting word (e.g., ``after", ``when"), which may then be specified by a proposition.  Similes and other comparisons by degree also behave in this way, but their connection is between the quality applied to each of the propositions.  Cause and effect-related concepts (e.g., because, since, if...then) usually connect two propositions, one being the cause and one being the effect.  Contrasts  also connect two propositions and specify them in the same way.  Some, like ``however", tend to have a clear orientation like cause and effect.  Others have a very similar meaning regardless of which proposition specifies the other.  It could be argued that there is always an orientation since the order in the surface structure emphasizes one proposition over the other.  If there is no difference in meaning, it would be more canonical to represent them as a group.  Conjunctions like ``and" and ``or" group together two or more concepts or propositions and are represented as, ``and $>$ [proposition 1, proposition 2, ... , proposition n]".   The reader may wonder how this example using ``and" differs from merely specifying a concept by a collection of propositions (or concepts).  The latter applies all of the propositions or concepts at the same time (e.g., The feathers were green-blue-yellow) whereas the former distinguishes each from the others (The feathers were green, blue, and yellow.)  

Our representation can also be used to organize the meaning of propositions between multiple sentences.  Sometimes, the connection is made explicitly by a word such as in the ``therefore"  at the beginning of a sentence.  In other cases, it may be less obvious.  Again, certain propositions will specify further details or explain aspects of another proposition and the connections in the network can be made accordingly.  A common discourse feature is quotation.  The quoted text becomes a semantic representation that specifies the theme of the speech act.  Another common discourse-level connector is sequence.  For example, ``She was frightened.  She ran."  In this example the first sentence leads the way to the actions in the second.  This could be represented as, (frighten $>$ [\{past\}, \{agent\} $>$ she]) $>$ \{seq\} $>$ (run $>$ [\{past\}, \{agent\} $>$ she]).  Unless made explicit by concepts in the sentences themselves, these discourse-level connections tend to be implied or inferred and thus will not normally be part of a first pass of converting surface text into a semantic representation.  


\section{Natural Language Generation (Realisation) and Parsing}
\label{sec-genparse}
While our representation stands on its own as a way of understanding meaning, adding the ability to generate surface text from it and parse surface text makes it much more practical.  In the following, we present an approach to doing so.  Please note that we provide only the briefest of treatments here, the details of which would fill a subsequent paper.

So far, we have introduced our representation and how to define concepts in terms of other concepts.  We now introduce the notion of ``rules", which relate a semantic network (or subnetwork) to the ordering of its constituent concepts in surface text.  Before we get started, however, we will say that whenever a semantic network is composed of a single concept alone, it is realised as the text of its label. For example, the concept labeled \emph{trust} will be generated as `trust'.  

Our simplest type of rule breaks a two-concept semantic representation into a sequence of two parts,

\begin{equation}
trust > \{past\} <=> [trust, `+ed']
\end{equation}

Here, we see that to express the past tense of the verb ``trust", the rule tells us that the trust concept will be followed by ``+ed" in the surface text.  And, since we know how to express the trust concept alone, we can substitute it in to get [`trust',`+ed'] or ``trusted".  Note that without the ``+" in ``+ed", we would have added a space to get ``trust ed".  It's also possible to use the same approach for prefixes (e.g., `un+', `wanted' = `unwanted' ) and to ``remove" letters (e.g., `berry', `-y', `+ies' = `berries'').  Although this simple example of generation (``trusted'')  is morphological in nature, the same mechanism is used to generate syntax as well.  Consider the more complex rule,

\begin{eqnarray}
\nonumber trust > [\{past\}, \{agent\} > he, \{theme\} > John]  \\ 
<=> [he, trust > \{past\}, John] 
\end{eqnarray}

This rule defines a subject-verb-object concept ordering commonly used by English propositions.  Using the total set of rules, we get the following sequence that generates surface text from a meaning representation.
\begin{eqnarray}
trust > [\{past\}, \{agent\} > he, \{theme\} > John] \\ \nonumber
[he, trust > \{past\}, John] \\ \nonumber
[`he', trust, `+ed', `John']\\ \nonumber
[`he', `trust', `+ed', `John']\\ \nonumber
[`he\,\, trusted\,\, John']
\end{eqnarray}

Since it is not practical to make rules for every combination of possible concepts, we will use similarity between the concepts' definitions to decide on-the-fly the likelihood of each rule being appropriate for use.  We can use our representation to build a lexicon and in it define,

\begin{eqnarray}
trust = \{verb\} \\
jump = \{verb\}
\end{eqnarray}

Then, if given the semantic representation

\begin{equation}
jump > \{past\},
\end{equation}

we can use the fact that both the trust and jump concepts are verbs and apply the ``trust''-associated rules in an analagous way,
\begin{eqnarray}
[jump > \{past\}] \\ \nonumber
[jump, `+ed'] \\\nonumber
[`jump', `+ed'] \\\nonumber
[`jumped']
\end{eqnarray}

Similarity between two concepts or subnetworks can be based on more than just concept definition.  The algorithm could compare the subnetwork shapes or use an ontology such as Wordnet \cite{Miller_1995} to involve synonomy and hypernonomy between concepts.  Another possibility is, with sufficient examples, to build up a \emph{concept}-to-vector representation like Word2vec \cite{MikolovEtAl_2013} or fastText \cite{BojanowskiEtAl_2016} and use those concept vector values to enhance the similarity metric.  

Parsing is generation in reverse.  From the surface text, we first seek an ordered set of concepts that when realised produces the full string.  Using the same rules, but in the reverse direction, we iteratively build subnetworks and networks of subnetworks that eventually become a single, complete semantic representation.  However, it is not expected to be as robust as generation since, like any parser, word sense disambiguation and implicit concepts are involved.  


With the rules we have provided in our examples, there was only a single surface structure or ``hypothesis" that could be generated.  As we add more rules and invoke the similarity metric, we will have multiple overlapping rules for transforming networks into surface structures.  When this happens, we duplicate the current hypotheses to separate the use of the overlapping rules.  Using the most similar rules, we can generate a variety of possible surface structures, which we can rank according to a similarity score derived from the rule similarities, a priori probabilities, sentence length, or other metrics.  For semantic networks with multiple valid surface texts, the valid surface texts will have the highest rankings. 


Finally, translating between two languages is the sequence of parsing the source surface text, transferring the source language concept network into a receptor language network, and expressing the receptor network as surface text.  The transfer step simply involves the replacement of concepts and subnetworks in the source language semantic network with concepts and subnetworks from the receptor language.  For this, a set of  ``transfer rules" must be defined.  As in the building of surface text from semantic representations, the transfer process is iterative and bifurcates with each new applicable transfer rule.

\section{Related Work}
\label{sec-relatedwork}
A wide variety of semantic representations exist, for a wide range of applications.  Here we discuss only a few in detail--those that are designed for language translation.  All three approaches we look at here are thematic-role-based representations.  According to Schubert \cite{Schubert_2015}, these ``...may prove to be of significant value in machine translation (since intuitively roles involved in a given sentence should be fairly invariant across languages).'' 

\subsection{Thematic-role-based Approaches}
The Translator's Assistant~\cite{BealeEtAl_2005}, \cite{Allman_2010} (TTA), begun in 2005, is, ``... a multilingual natural language generator based on linguistic universals, typologies, and primitives".   Focusing on providing an interface suited to linguists, the purpose is to provide draft translations that will allow human translators to complete translations more efficiently.  In this system, manual effort is required to generate the desired interlingual representation of a source text and to build up the natural language generation rules for new languages.  Then, authors that want to translate their works will encode them once and generate them in the available languages.  Allman~\cite{Allman_2010} indicates that translators' speeds increased four-fold with the use of this approach.  Work on this approach continues, although mostly among the originators.  

The Universal Networking Language \cite{UchidaEtAl_2006} or UNL is a language-independent knowledge base representation introduced by the United Nations University in 1996.  Later a separate organization was formed to support its development.  Primarily UNL serves as an interlingua into which information can be encoded and later generated into multiple languages.  Substantial work has gone into the development of web tools for authoring UNL representations and adding to the grammars and lexicons of a variety of languages.  There are also tools for receptor language realisation, but their official ``text-to-text" initiative for automated translation has not yet been released.  A few parsers called ``enconverters" and multiple realisers called ``deconverters" have been developed by various parties for transduction between specific languages and UNL. From searching the main technical website~\cite{unlwebsite_2017}, it seems like little work on this approach has occurred since 2015, although a few papers have been published since then that use UNL.

KANT \cite{MitamuraEtAl_1991}, which is derived from ``Knowledge-based, Accurate Natural-language Translation" was introduced in 1989 at the Center for Machine Translation at Carnegie Mellon University.  It was used to translate technical English into French, German, and Spanish.  Translators would write their technical English into a specialized editor that would help them constrain their expressions to those that could be correctly translated by the system.  The technical English would be automatically interpreted into a standardized interlingua, which was finally used to generate text in the receptor language.  In essence, this system could actually translate English text into several other languages automatically.  Although still available, development was discontinued in the early 2000's.


\subsection{The Semantic Representations}
\begin{figure*}[h]
\centering
\includegraphics[width=\textwidth]{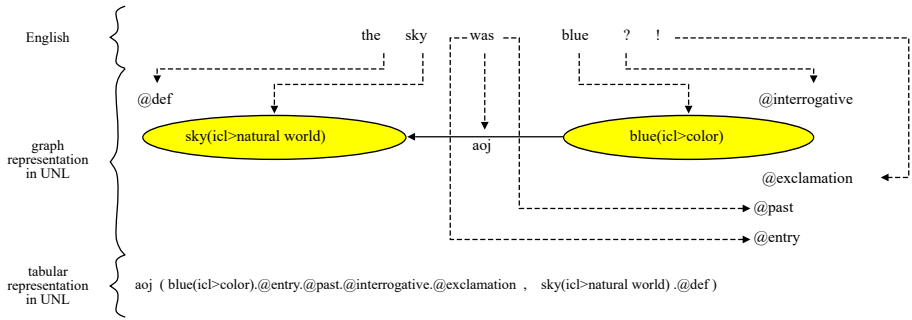}
\caption{Example of a UNL graph for ``The sky was blue?!" }
\label{fig:UNLgraph}
\end{figure*}

The semantic representations of these approaches are similar in some ways and quite different in others.  UNL provides a summary of the types of features contained within a thematic-role-based semantic representation.  Figure~\ref{fig:UNLgraph} \cite{unlgraph_2017} captures the main elements called ``Words", ``Attributes", and ``Relations".  Words are simply concepts.  Attributes are set of concepts that provide additional information (e.g., tense, gender, manner, lexical category, etc.).  There are more than 300 officially defined attributes falling into 33 categories and subcategories.  Relations describe the way in which two Words interact.  Relations, sometimes referred to as case roles, are directed and can take any of the 40 predefined values in UNL.  Like our approach, UNL also has a mechanism for encapsulating subgraphs within a Word, allowing for high-level Relations.  

TTA defines their semantic representation a little differently.  Concepts fall into one of seven categories, most notable of which are the Object (i.e., noun), Event (i.e., verb), and Relation (i.e., case role) concepts.  Object concepts have 6 ``features" each (e.g., polarity, proximity, number).  Each feature is set to one of a number of mutually exclusive values that have been predefined according to the feature type.  Events have four features a piece (time, aspect, mood, and polarity), which similarly have many possible predefined settings.  There are nine predefined Relations that can be used to relate Objects and Events.  Propositions are attributed with a further 15 different features (e.g., type, discourse genre, speaker), each having their own set of values.  In total, there are more than 250 possible feature values that are used in this approach.  

KANT represents concepts as ``frames" that are somewhat like functions with ``slots" or function arguments.  Frames represent objects, events, and properties, each having a certain number of slots.  The slot values can be limited to certain classes of concepts, according to the associated ontology. This helps it to do a better job of disambiguation during parsing, by finding the concept sense with constrained slots that are most suited to the other parsed inputs.  In the case of an Event frame (i.e., a verb), the agent and theme are key case roles that are to be filled.  A sentence can take the form of an instantiated event frame that specifies the values of its necessary slots and can add features such as necessity, mood, and voice.  The expressiveness of the KANT system is limited to its vocabulary and use of technical English, in spite of being a multilingual translation system.

In summary, all three approaches invoke a complex set of features that must be learned by prospective users when performing generation or when building grammars for additional languages.  This makes the learning curve steep, setting up a barrier to entry.  UNL offers free online training to achieve several certifications to help mitigate this.  In all approaches, the building of language rules is almost entirely manual, complicated, and very time-consuming.  

The existing approaches interpret or label features of a sentence with technical linguistic or ontological terms, whereas the current approach allows these to be abstracted away and presents commonly understood concepts on the surface.  This makes the semantic representation much more readable to the layman and therefore more accessible, especially if the terms are in a different language than the one being worked in.  It also means that there are fewer key concepts to remember.  Also in our approach, by forcing every bit of meaning to become a concept and reducing the use of traditional case roles, a user or automated system does not have to decide whether the bit of meaning should take the form of a case role, a concept, or a property.  There is less interpretation required overall for human and machine alike. In total, these benefits will make all aspects of language building more efficient, even if done manually.

\subsection{Generation}
KANT generates surface text in the receptor language by first finding receptor concepts to match each of the interlingual concepts, and then forms a receptor language ``f-structure" or describes it in terms of suitable grammatical elements present in the receptor language.  Finally, the concepts are inflected and ordered according to the receptor language rules.  Generation rules are akin to functions.  Embedded in a rule is a set of concept qualities that are compared against a to-be-generated interlingual representation.  If there is a match, the rule is applied, grammatically organizing parts of the proposition (the f-structure).

UNL uses a complex system of rules to generate receptor text.  There are different types of ``transformation rules" (T-rules) that transform networks into simpler networks, networks into trees, trees into lists of concepts, and lists into surface text.  The various rules describe the kinds of functional transformations that their associated structures can undergo (e.g., add, replace, or delete concepts, etc.).  The receptor language-specific rules are defined in terms of these T-rules.  The receptor language generation rules are ordered giving earlier rules priority in generation. The rules are applied recursively, that is, repeatedly on the resulting outputs, until they no longer apply, and then move onto the next T-rule in the grammar.  The definition of a T-rule is lengthy, though we will not go into detail here.  

Like KANT, TTA uses a two-stage approach to generating text.  The semantic representation is converted to a receptor-language structure using the ``transfer grammar" and ultimately into receptor-language text using a ``synthesizing grammar".  It is no easy task to convert from one collection of concepts to another because while there will be many direct concept matches between two languages (or an interlingua and receptor language), differences in culture will mean the absence of certain concepts and linguistic features in one of the languages.  TTA's transfer grammar employs a sequence of nine different kinds of transfer rules.  Of these, the most complicated would appear to those that require some form of deduction or knowledge of the wider context.  Necessary features in the receptor language that are not present in the source language seem particularly challenging.  In the case of honorifics, the relationship between speakers in a dialog must be determined from the context.  Although not discussed here because TTA does not parse, pronouns must have their referent identified for proper translation into languages that do not use pronouns, or that use pronouns differently.  Also, when the receptor language requires the explication of a concept in other terms, and that concept occurs frequently in a short span, repeated explication can make the translation unnatural and misleading.  TTA solves this problem by providing the explication the first time the concept is used and then a referring concept (behaving like a pronoun) thereafter.  According to Allman~\cite{Allman_2010}, the building of a language's transfer grammar is more time-consuming than building the synthesizing grammar.  In TTA's synthesizing grammar, there are eight types of rules.  These include spellout rules, clitic rules, phrase structure rules, and more.  Of these, the most complicated would seem to be the choice of when to use pronouns and when not to use them.  At any rate, it would seem that keeping track of which reified entity each pronoun or referring concept points to will be necessary for accurate translations.
 

It would appear that none of these three approaches are thriving.  Perhaps it is because there is 1) limited application, 2) the learning curve is too steep, and 3) the manual effort required is more than the perceived benefit.  Users are willing to trial approaches that are simple to use and quick to get useful results.  In contrast, approaches that require a lot of manual effort and skill training and still require manual editing afterward are less appealing.  Our approach aims to lessen the impact of these factors, yet is not without its challenges.  To name a few: 
\begin{itemize}
\item learning the list of stemless concepts.  Although we try to minimize the learning curve associated with this approach, there will be a number of stemless concepts to define and/or learn, especially for transfer from a specific language semantic representation to the interlingua, which will have the broadest set of stemless concepts.  
\item complicated structures.  Representing certain syntactic phenomena (e.g., relative clause) can be complicated.  As in the \{have\} concept example, it would be possible to define some of these to make them easier to grasp and use.
\item language design consistency.  Because we put so much ``power'' in the hands of the linguist to shape the grammar, lexicon, and indeed the semantic representation itself, variability will occur.  This will be greatest when the user is dealing directly with semantic networks in the receptor language and least when dealing with a common interlingua, to which all receptor language definitions would ultimately interface.
\end{itemize}

\section{Conclusion}
Developing a semantic representation that is versatile yet comprehensive is a challenging task, especially for the purpose of language translation which demands both.  The present semantic representation seeks to do this by providing a minimalist structure based on specification, breaking down meaning at the morpheme level, and employing encapsulation which serves to assert canonicity, support high-level connections between propositions, and define concepts hierarchically.  It is a departure from standard thematic-role-based representations that have been designed with language translation in mind.  It is more accessible to linguists and software developers because it largely strips away case roles and abstracts away many underlying linguistic and ontological details.  It also naturally allows for data-driven methods to be built on top of it.  Such novel qualities will be necessary to enable RBMT processing to stand alongside big data machine translation approaches in their respective areas of usefulness.  Much work remains but, if successful, such an approach has the potential to enable a greater coverage of the world's languages for automated or semi-automated translation and other useful NLP tasks.


\bibliographystyle{spphys}       
\bibliography{paper}   

\end{document}